\documentclass[5p,times,twocolumn, authoryear]{elsarticle}
\usepackage{amsmath,amssymb,amsfonts,color}
\usepackage{graphicx}
\usepackage{textcomp}
\usepackage[dvipsnames]{xcolor}
\usepackage{enumerate}
\usepackage{setspace}
\usepackage{bm}
\usepackage{comment}
\usepackage{caption}
\usepackage{tipa}
\usepackage{footnote}
\usepackage{subfigure}
\usepackage{algorithm}
\usepackage{algorithmic}
\usepackage{marvosym}

\usepackage{multirow}
\usepackage{makecell}
\usepackage{url}
\usepackage{booktabs}
\usepackage{tabularx}
\usepackage{adjustbox}
\usepackage{arydshln}

\def\x{{\mathbf x}}

\usepackage{hyperref}

\newcolumntype{C}[1]{>{\centering\arraybackslash\vspace{0pt}}m{#1}}
\newcolumntype{Y}{>{\centering\arraybackslash}X}
\newcolumntype{P}{>{\centering\arraybackslash}p}

\begin{document}

\begin{frontmatter}
\title{Foreground-Aware Dataset Distillation via Dual-Path Patch Selection}

\author[a]{Mingzhuo Li}
\ead{mingzhuo@lmd.ist.hokudai.ac.jp}
\author[a]{Guang Li}
\ead{guang@lmd.ist.hokudai.ac.jp}
\author[a]{Longzhen Li}
\ead{longzhen@lmd.ist.hokudai.ac.jp}
\author[a]{Ren Togo}
\ead{togo@lmd.ist.hokudai.ac.jp}
\author[a]{Keisuke Maeda}
\ead{maeda@lmd.ist.hokudai.ac.jp}
\author[a]{Takahiro Ogawa}
\ead{ogawa@lmd.ist.hokudai.ac.jp}
\author[a]{Miki Haseyama\corref{cor1}}
\ead{mhaseyama@lmd.ist.hokudai.ac.jp}
\cortext[cor1]{Corresponding author}

\address[a]{Hokkaido University, Japan}



\begin{abstract}
In this paper, we propose a foreground-aware dataset distillation method that performs content-adaptive dual-path patch selection. With the rising computational cost of training large-scale deep models, dataset distillation has emerged as a promising approach for constructing compact synthetic datasets that retain knowledge from large original datasets. Although optimization-based methods can achieve strong performance, they often incur considerable computational and memory costs and may produce synthetic images that generalize poorly across architectures. Recent non-optimization-based methods alleviate these limitations by constructing distilled images from real-image patches. However, their inflexible patch-construction strategies may either retain redundant background content or discard important foreground structures. To address this issue, we propose a foreground-aware dataset distillation framework with content-adaptive dual-path patch selection. We first use Grounded SAM2 to estimate the foreground occupancy of each image and derive category-specific routing thresholds from the corresponding class-wise distributions. Each image is routed to one of two complementary paths: the candidate-crop selection path for images with relatively sparse foregrounds, or the direct transformation path for images with relatively dominant foregrounds. This dual-path mechanism balances foreground concentration and preservation, reducing redundant background content while retaining the structural integrity of foreground objects at the dataset level. Experiments on CIFAR-10, CIFAR-100, ImageNette, and ImageWoof with two network architectures and three distillation budgets demonstrate that the proposed method outperforms the baseline method RDED in nearly all evaluated settings. Further analyses on hyperparameters, transformation strategy, routing strategy, and time cost further demonstrate the effectiveness of foreground-aware dual-path patch selection for efficient dataset distillation.

\end{abstract}


\end{frontmatter}
\section{Introduction}
Driven by rapid advances in computing hardware, deep learning has made remarkable progress in recent years~\citep{schmidhuber2015deep, matsuo2022deep}. Large-scale models such as BERT~\citep{devlin2019bert}, Stable Diffusion~\citep{rombach2022high}, and ChatGPT~\citep{achiam2023gpt} have achieved impressive performance across a broad range of tasks. However, this progress also leads to a sharp increase in training costs~\citep{dargan2020survey, alzubaidi2021review}. Reducing the computational burden of training large models and limiting excessive resource consumption have become central concerns in modern AI research. Among the approaches developed to address these concerns, dataset distillation has emerged as an attractive research direction~\citep{li2022awesome, yu2023review, liu2025survey}, which has been applied across various fields, including privacy protection~\citep{dong2022privacy, li2020soft, li2022compressed}, graph neural networks~\citep{jin2022graph, jin2022condensing, liu2024gcsr}, federated learning~\citep{liu2023meta, wang2024fed, jia2024feddg}, and reinforcement learning~\citep{lupu2024bd, lei2024obl, wilhelm2025rl}.

\par
The pioneering work of ~\citet{wang2018datasetdistillation} formulated dataset distillation as a meta-learning problem and defined the goal as synthesizing a compact dataset that can train a model to achieve downstream performance comparable to that obtained using the original dataset. This formulation naturally leads to a bi-level optimization problem, which is computationally demanding in practice. To reduce this cost, subsequent studies have explored more efficient single-level optimization strategies that can be broadly grouped into three representative categories. First, gradient-matching methods constrain synthetic data to produce gradients similar to those obtained from real batches~\citep{zhao2021gm, zhao2021differentiatble, zhang2025gradient}. Second, training-trajectory-matching methods extend this principle by matching the full training trajectories obtained using the original dataset instead of individual gradient steps~\citep{cazenavette2022dataset, du2023minimizing, guo2024datm}. Third, feature- or distribution-matching methods align the representations of real and synthetic data, typically using features extracted by a pretrained network~\citep{zhao2023dm, wang2022cafe, li2025hdd}.

\par

Despite these advances, conventional optimization-based dataset distillation methods still face several bottlenecks. 
First, many methods remain computationally expensive, particularly those based on bi-level optimization, which requires substantial computational resources and memory~\citep{cui2023scaling}. The computational requirement limits efficiency and hinders practical deployment on large-scale or high-resolution datasets. Second, optimization-based approaches often produce unrealistic, noise-like synthetic images and may overfit to the architecture used during synthesis, weakening cross-architecture generalization~\citep{cazenavette2023generalizing}. Moreover, the resulting synthetic dataset may capture only a part of the various visual patterns from the original dataset depending on the optimization objective, especially under a highly constrained distillation budget.

\par

In contrast, coreset methods provide an optimization-free and real-data-based alternative by selecting samples from the original dataset. Representative approaches include random selection, Herding~\citep{Welling2009Herding}, K-Center~\citep{sener2017kcenter}, example forgetting~\citep{toneva2019empirical}, and the EL2N score~\citep{paul2021deep}. Their selection criteria capture different notions of representativeness, feature-space coverage, or sample difficulty, allowing the selected subset to retain a certain degree of information diversity while preserving realistic visual patterns. Nevertheless, these methods operate at the level of complete images. Under an extremely limited data budget, redundant background or contextual content in the selected images may consume a disproportionate share of the available pixel budget, while potentially informative regions in unselected images are discarded.

\par

These limitations have motivated non-optimization dataset distillation paradigms, including generative~\citep{su2024d4m, gu2024efficient, su2024diffusion}, decoupled~\citep{yin2023sre2l, shao2024generalized, yin2024dataset}, and selection-based approaches~\citep{ye2025sel, lee2024selmatch}. Generative and decoupled methods avoid directly optimizing synthetic pixels by reconstructing or generating distilled data with learned generators or reconstruction networks. In contrast, selection-based approaches operate directly on real-image content and construct distilled datasets by selecting and recombining informative regions from the original data, increasing information density while maintaining visual realism. A representative example is the RDED proposed by \citet{sun2024rded}, which introduces the realism score and aggregates real-image patches with high scores into distilled samples. By reusing real patches instead of optimizing synthetic pixels, selection-based methods avoid costly synthesis loops, improve computational efficiency, and often yield distilled data with stronger realism and cross-architecture transferability.

\par

\begin{figure}[t]
        \centering
        \includegraphics[width=\linewidth]{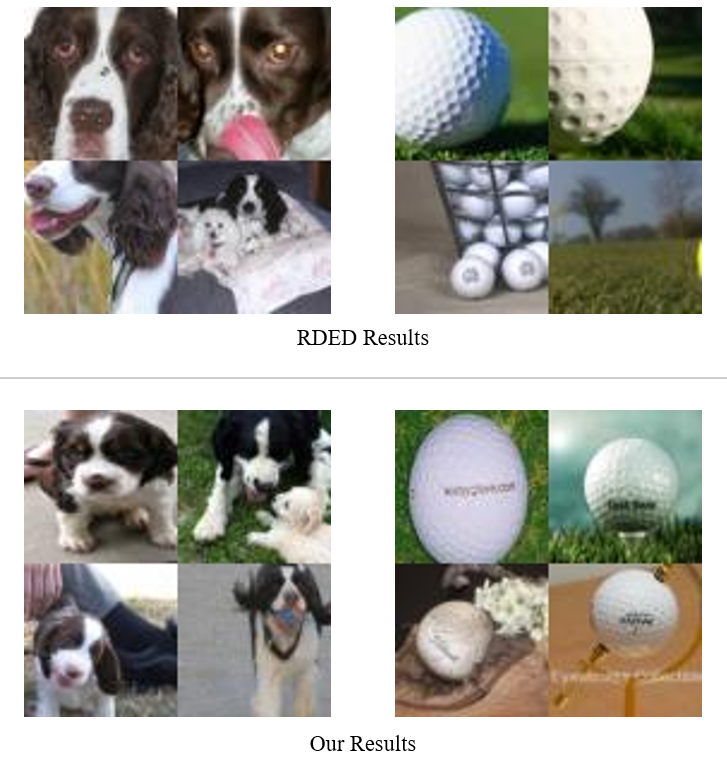}
        \caption{Comparison of images generated by RDED and the proposed method. RDED may include patches with partial foreground objects, while the proposed method shows better foreground preservation.}
        \label{fig:intro}
\end{figure}

Despite these advantages, existing selection-based distillation techniques share a fundamental limitation: they generally apply fixed patch-extraction and selection rules without considering the underlying image structure. Real-world datasets, however, contain categories with widely varying spatial layouts and foreground-background distributions. When the foreground occupies a large portion of an image or has an irregular geometry, fixed extraction rules may fail to preserve the complete semantics of the primary object. As illustrated in Fig.~\ref{fig:intro}, RDED may extract incomplete or uninformative patches, discard crucial visual cues, and ultimately reduce the downstream accuracy.

\par
Motivated by this observation, we reconsider selection-based dataset distillation from a foreground-aware perspective and develop a content-adaptive dual-path patch-selection framework. Our approach extends the RDED pipeline with two core components: foreground-aware preprocessing and content-adaptive routing. We first employ Grounded SAM2~\citep{ravi2024sam2, ren2024grounded, jiang2024trex2} to identify the foreground region of each image and compute its foreground occupancy ratio. We then use the category-wise distributions to derive class-specific routing thresholds. Based on these thresholds, the proposed module determines whether to select an informative crop or directly transform the source image into the selected patch. The candidate-crop selection path improves foreground concentration for images with relatively sparse foregrounds, whereas the direct transformation path preserves foreground structures for images with relatively dominant foregrounds. In this way, the distilled images preserve more complete foreground information while reducing redundant background content. At the same time, content-adaptive routing mitigates the risk of discarding foreground parts, especially when the foreground object is difficult to capture within randomly cropped patches.

\par

The main contributions of this work are summarized as follows:

\begin{itemize}
\item We introduce foreground occupancy ratios as lightweight structural cues and derive category-specific routing thresholds to judge the relative dominance of the foreground object.
\item We propose a foreground-aware dual-path patch-selection strategy that balances foreground concentration and preservation by routing images to the candidate-crop selection path or the direct transformation path.
\item Extensive experiments demonstrate consistent improvements over the baseline method across multiple datasets, distillation budgets, and model architectures.
\end{itemize}

\section{Related Work}

\subsection{Optimization-based Dataset Distillation}
Optimization-based dataset distillation synthesizes a compact set of samples that reproduces informative signals from the original dataset. Gradient-matching methods align gradients from real and distilled data at individual optimization steps~\citep{zhao2021gm, zhao2021differentiatble, lee2022dataset, zhang2025gradient}. Trajectory-matching methods align longer training dynamics or higher-order properties of expert models~\citep{cazenavette2022dataset, du2023minimizing, shin2023lcmat, du2023seqmatch, guo2024datm}. Although these formulations have improved distillation quality and enabled larger-scale experiments, repeated optimization over synthetic samples or expert trajectories can still be costly~\citep{cui2023scaling}.

\par

A parallel line of work aligns real and synthetic data in representation space. For example, CAFE~\citep{wang2022cafe} matches features extracted by a pretrained network, whereas DM and IDM explicitly reduce feature-distribution discrepancies~\citep{zhao2023dm, zhao2023idm}. Subsequent methods incorporate attention relations, sample-feature dependencies, or latent distribution statistics to preserve richer structure for alignment~\citep{sajedi2023datadam, deng2024iid, wei2024lqm, zhang2024m3d}. Although these methods reduce some limitations of the aforementioned methods, they generally retain an optimization-dependent synthesis procedure.

\subsection{Non-optimization and Selection-based Distillation}
Non-optimization approaches avoid directly optimizing synthetic pixels. Generative and decoupled methods reconstruct or generate distilled data using pretrained networks or generators~\citep{li2024generative, gu2024efficient, su2024d4m, su2024diffusion, yin2023sre2l, yin2024dataset, shao2024generalized}. These approaches can reduce the cost of bi-level optimization and improve the visual realism of distilled examples.

\par

Selection-based distillation instead reuses content from the original images. Among them, RDED~\citep{sun2024rded} is a representative method. It ranks randomly sampled patches using a realism score and aggregates selected patches to form distilled images. This formulation efficiently preserves realistic image content, but fixed crop selection does not account for the foreground scale or spatial layout of individual images and categories. Recent methods explore complementary directions, including progressive real-sample selection~\citep{chen2025ccfs}, critical-region perspective~\citep{tran2025nrrdd}, and the integration of informative local patches with global views~\citep{hu2026focusdc}. In contrast to these methods, our work focuses on the crop-selection paradigm and introduces a foreground-aware dual-path patch selection to obtain informative patches in a content-adaptive way.

\begin{figure*}[t]
        \centering
        \includegraphics[width=18cm]{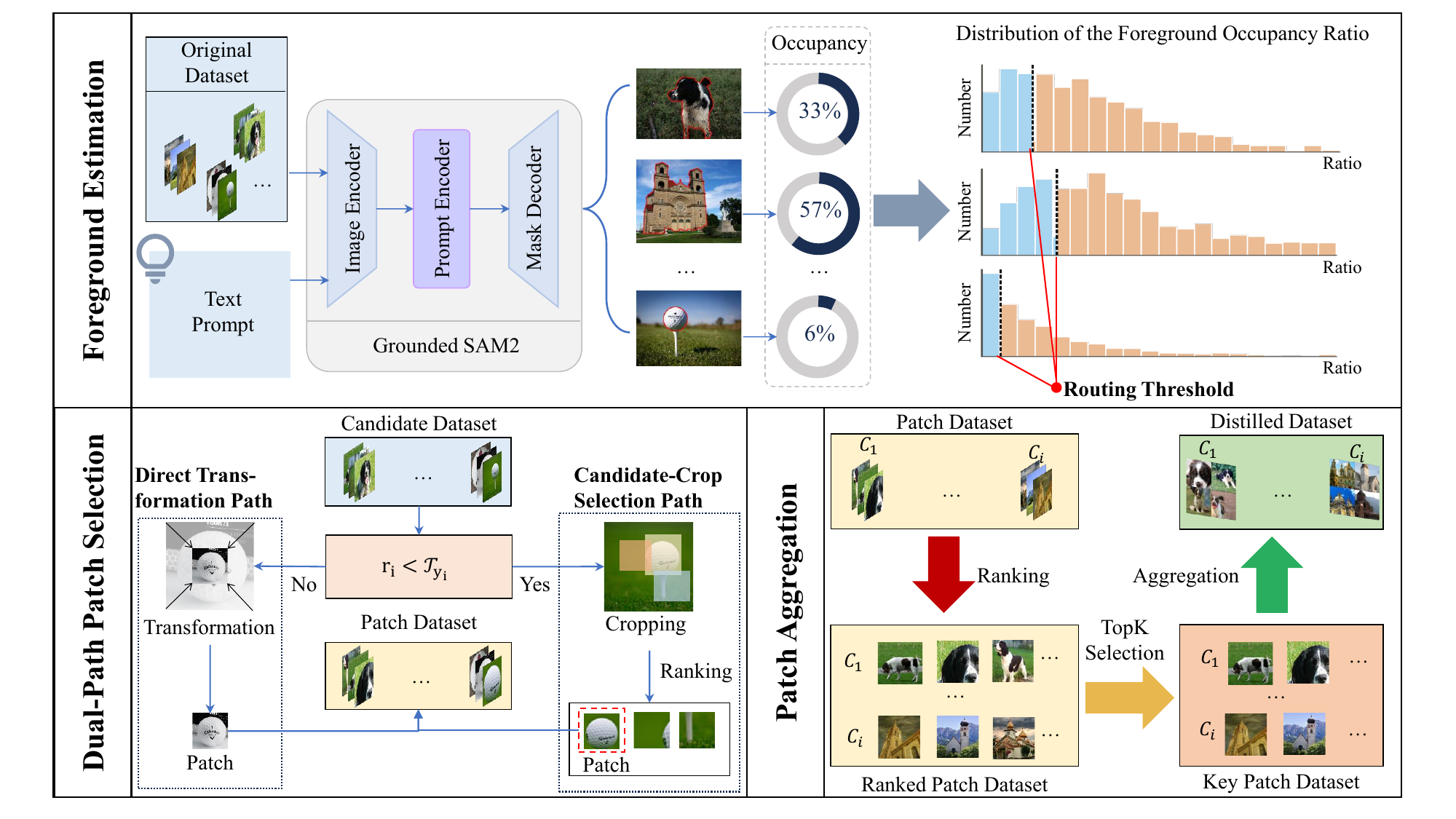}
        \caption{
        Overview of the proposed foreground-aware dual-path patch selection methodology. In the Foreground Estimation stage, Grounded SAM2 is applied to all images to obtain foreground occupancy ratios. The category-specific routing thresholds are computed from the ratio distributions of each class. In the dual-path patch Selection stage, a candidate subset is sampled uniformly from the original dataset. The images are routed to the candidate-crop selection path or the direct transformation path based on the routing threshold. In the patch aggregation stage, top-$k$ patches are selected, where $k$ depends on the distillation budget, and then sequentially aggregated into distilled images.
        }
        \label{fig:workflow}
\end{figure*}

\section{Methodology}
We first use Grounded SAM2 to estimate the foreground occupancy ratio of the original images and the category-wise distribution. Then, inspired by RDED, we construct a class-balanced candidate subset and obtain the patch dataset by performing the proposed dual-path patch selection. The patch dataset is then aggregated into the distilled dataset. The method aims to retain task-relevant foreground information while removing redundant background by routing each candidate image according to its foreground occupancy ratio. As shown in Fig.~\ref{fig:workflow}, the processing can be divided into three stages: foreground estimation, dual-path patch selection, and patch aggregation.

\subsection{Foreground Estimation}
Let $\mathcal{D}=\{(\bm{x}_i,y_i)\}_{i=1}^{N}$ denote the original dataset, where $y_i\in\mathcal{C}=\{1,\ldots,C\}$ is the class label of image $\bm{x}_i$. For each class $c$, we specify a text prompt $l_c$ that describes its foreground object. For every image in $\mathcal{D}$, Grounded SAM2 is used to obtain a binary foreground mask $F_i$ as follows:
\begin{equation}
F_i=G_{\mathrm{GSAM2}}(\bm{x}_i,l_{y_i}), \qquad F_i\in\{0,1\}^{H_i\times W_i} \cup\{\emptyset\},
\label{eq:foreground_mask}
\end{equation}
where $G_{\mathrm{GSAM2}}(\cdot)$ denotes Grounded SAM2 and $H_i$ and $W_i$ are the height and width of $\bm{x}_i$, respectively. For some images, Grounded SAM2 may fail to detect a foreground object and therefore return no valid mask, which we denote by $F_i=\emptyset$. To maintain a unified definition of the occupancy variable, we assign $r_i=0$ to these images. For images with valid masks, the occupancy ratio is calculated from the proportion of foreground pixels. Accordingly, $r_i$ is defined as follows:
\begin{equation}
r_i=
\begin{cases}
0, & F_i=\emptyset \\[3pt]
\frac{1}{H_iW_i}\sum_{u=1}^{H_i}\sum_{v=1}^{W_i}F_i(u,v), & F_i \ne \emptyset.
\end{cases}
\label{eq:foreground_occupancy}
\end{equation}

\par

Thus, $r_i > 0$ measures the spatial prominence of the foreground in image $\bm{x}_i$. From the perspective of selection-based dataset distillation, a useful distilled sample should retain recognizable, class-relevant evidence within a limited pixel budget. A low occupancy ratio indicates that only a small fraction of the pixels in $\bm{x}_i$ belong to the estimated foreground, while the remaining pixels predominantly represent background or other contextual content. In this case, cropping can improve foreground concentration by identifying a region with stronger class-relevant evidence. Conversely, a high occupancy ratio indicates that the foreground already occupies a substantial portion of the image and provides abundant task-relevant visual content. Further cropping is more likely to truncate important object parts or their spatial relationships.

\par

However, foreground occupancy alone does not provide an exact geometric measure of an image's suitability for cropping. Images with the same occupancy may contain objects with different shapes, aspect ratios, and spatial extents. For example, a thin object that spans an image may be truncated even when its occupancy is relatively low. We therefore use the occupancy ratio as a category-relative routing cue rather than a universal absolute criterion for cropping. Specifically, an image is not routed to a crop candidate merely because its occupancy ratio falls below a predefined threshold. Instead, for each category, we consider the empirical distribution of foreground occupancy ratios across the images and derive a category-specific threshold at a fixed quantile. In this way, the corresponding thresholds represent the underlying occupancy preferences within each category, facilitating an assessment of an image's category-wise relative dominance. Images with relatively sparse foregrounds are routed to the candidate-crop selection path, while the remaining images are processed by the direct transformation path. Specifically, images without valid foreground masks are routed directly to the candidate-crop selection path, allowing patch selection to proceed without relying on foreground occupancy information.

\par

Let $\mathcal{D}_c=\{(\bm{x}_i,y_i)\in\mathcal{D}\mid y_i=c\}$ denote the images of class $c$, and $\mathcal{I}_c=\{i\mid (\bm{x}_i,y_i)\in\mathcal{D}_c\}$ be the index set of class $c$. For a fixed routing rate $\rho\in(0, 1)$, the category-specific threshold is defined as follows: 
\begin{equation}
\mathcal{T}_c=\operatorname{Quantile}_{\rho}\left(\{r_i\mid i\in\mathcal{I}_c , \
 F_i \ne \emptyset \} \right), 
\label{eq:category_threshold}
\end{equation}
where $\operatorname{Quantile}_{\rho}(\cdot)$ denotes the empirical
$\rho$-quantile operator. Images without valid foreground masks are excluded in the calculation to prevent bias in estimating category-wise relative foreground dominance resulting from foreground-detection failures.

\par

Calibrating the routing rate from the class-wise occupancy distribution implicitly reflects category-specific tendencies in object scale and layout. Consequently, $\mathcal{T}_c$ provides a relative reference for deciding whether cropping is likely to concentrate useful foreground content instead of guaranteeing that a crop containing the complete object exists.

\par

\subsection{Dual-Path Patch Selection}
As discussed by \citet{sun2024rded}, 
processing the entire dataset incurs substantial computational cost, while directly selecting the highest-scoring samples may introduce selection bias and reduce data diversity. We therefore follow the practical implementation of RDED and first construct a uniformly sampled candidate subset. For each class, we sample a subset $\mathcal{D}^{\mathrm{cand}}_c\subset\mathcal{D}_c$ containing $N_{\mathrm{cand}}$ images per class and form the candidate dataset $\mathcal{D}^{\mathrm{cand}}=\bigcup_{c=1}^{C}\mathcal{D}^{\mathrm{cand}}_c$. All subsequent operations are performed on the candidate dataset $\mathcal{D}^{\mathrm{cand}}$. 

\par

Each candidate image is routed to either the candidate-crop selection path or the direct transformation path by comparing its foreground occupancy ratio $r_i$ with the threshold $\mathcal{T}_{y_i}$ obtained from Eq.~(\ref{eq:category_threshold}).

\subsubsection{Candidate-Crop Selection Path}
For the candidate-crop selection path, a single random crop is sensitive to its location and may either miss the foreground object or retain a large amount of irrelevant background. For each image $\bm{x}_i$, we form a candidate patch set consisting of $K_\text{crop}$ randomly cropped and resized square patches to provide multiple spatial views, reducing the variance introduced by random cropping and improving robustness across foreground situations. The candidate patches are then resized to the target patch resolution $(h_p,w_p)$ for subsequent scoring and aggregation. The candidate patch set $\mathcal{P}_i$ is sampled as follows:
\begin{equation}
\mathcal{P}_i = \operatorname{Crop}(\bm{x}_i,K_\mathrm{crop}) = \{ \bm{p}_{i,k} \}_{k=1}^{K_\mathrm{crop}}, 
\label{eq:cropping}
\end{equation}
where $\operatorname{Crop}(\cdot)$ denotes the cropping process and $\bm{p}_{i,k}$ denotes the cropped patch.

\par

Following ~\citet{sun2024rded}, the realism score is estimated using a pretrained observer model. Because the patch size is always smaller than the input size, we add a preprocessing step to the model that pads each patch with zeros to the input size. With $\ell(\cdot,\cdot)$ denoting the cross-entropy loss, the realism score is calculated as follows:
\begin{equation}
S(\bm{p}_{i,k}) = - \ell \bigl( \phi_{\theta_T}(\bm{p}_{i,k}), y_i \bigr),
\end{equation}
where $\phi_{\theta_T}(\cdot)$ denotes the observer model with preprocessing. With this function, a patch receives a high score when the observer recognizes its class with high confidence. In this path, the selected patch $\bm{p}^{\text{crop}}_i$ is obtained as follows: 
\begin{equation}
\bm{p}^{\mathrm{crop}}_i = \underset{ \bm{p} \in \mathcal{P}_i } { \operatorname{arg \ max} } \ S(\bm{p}).
\label{eq:crop_selection}
\end{equation}

\subsubsection{Direct Transformation Path}
In the direct transformation path, a transformation operator converts the candidate image directly into the selected patch as follows:
\begin{equation}
\bm{p}^{\mathrm{trans}}_i = G(\bm{x}_i;(h_p,w_p)),
\label{eq:direct_transformation}
\end{equation}
where $G$ denotes the image transformation. We use \emph{Squash} as the default transformation, which resizes the complete image to $(h_p,w_p)$ without cropping and therefore preserves its full field of view. We also consider alternative transformations and compare their differences in Sec.~\ref{sec:transform}.

\subsubsection{Foreground-Aware Path Routing}
For image $\bm{x}_{i}$ with occupancy ratio $r_i$, the candidate-crop selection path is used when $r_i < \mathcal{T}_{y_i}$ because the foreground is relatively sparse within its category and cropping can remove redundant background. Otherwise, the foreground occupies a comparatively large portion of the image, making cropping more likely to truncate extended object structures. In this situation, the image is routed to the direct transformation path. 

\par

Accounting for images without valid foreground masks, the final patch $\bm{p}^{\star}_i$ is obtained by the dual-path selection as follows:
\begin{equation}
\bm{p}^{\star}_i=
\begin{cases}
\bm{p}^{\mathrm{crop}}_i, & F_i = \emptyset \ \lor \ r_i<\mathcal{T}_{y_i} \\[3pt]
\bm{p}^{\mathrm{trans}}_i, & F_i \ne \emptyset \ \land \ r_i\geq\mathcal{T}_{y_i}.
\end{cases}
\label{eq:dual_path_selection}
\end{equation}

\subsection{Patch Aggregation}
\begin{figure*}[t]
        \centering
        \includegraphics[width=18cm]{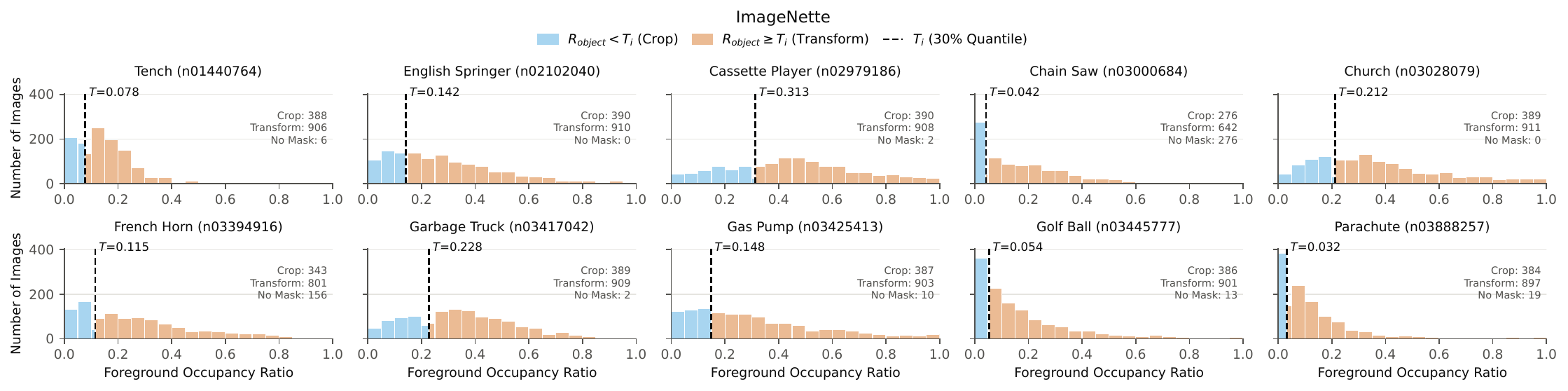}
        \includegraphics[width=18cm]{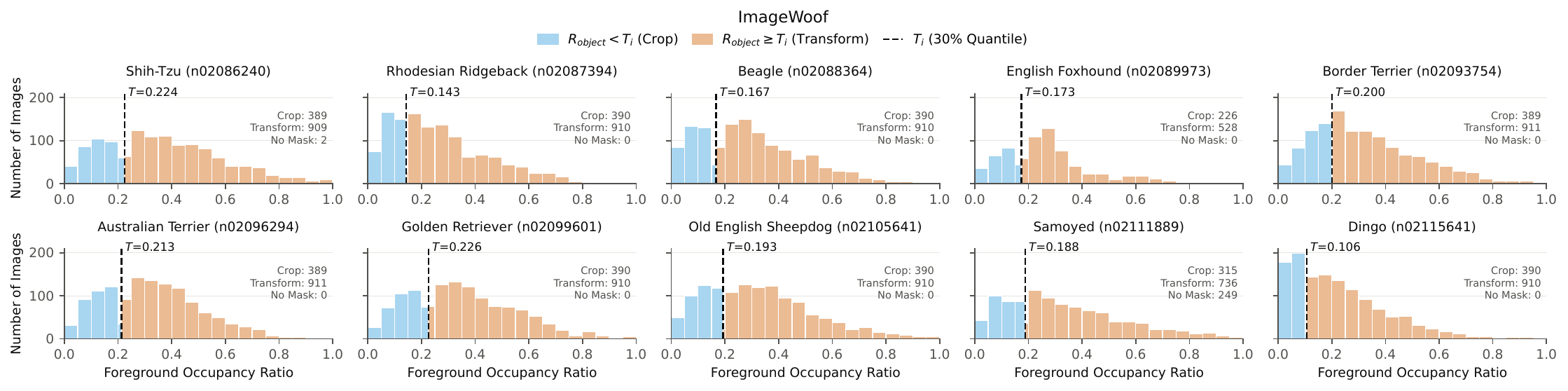}
        \includegraphics[width=18cm]{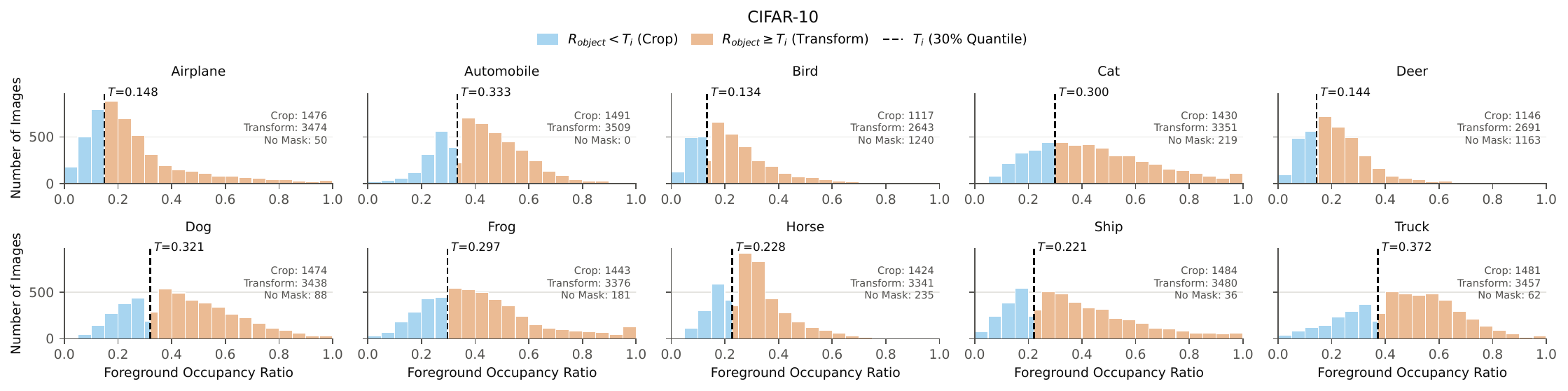}
        \caption{
        Per-class distributions of the foreground occupancy ratio $r_i$ on ImageNette, ImageWoof, and CIFAR-10. Each bar counts the images of a class whose $r_i$ falls within the corresponding bin. The dashed line marks the routing threshold $\mathcal{T}_c$. Images without valid foreground masks (No Mask) are excluded from the bars. Blue areas show remaining images that are routed to the candidate-crop selection path (Crop), and orange areas show those routed to the direct transformation path (Transform). Detailed image numbers are listed in the spare space.
        }
        \label{fig:distribution}
\end{figure*}

To combine the representative characteristics from several source images and increase the information density and diversity of the distilled dataset, we aggregate multiple patches into a single distilled image rather than using the patches as distilled images. The number of patches aggregated into one distilled image (patches per image, $Z$) controls a trade-off between information diversity and spatial fidelity. A small $Z$ limits the amount of class-relevant content in each distilled image, whereas a large $Z$ reduces the resolution of each constituent patch because of the fixed final image resolution. For implementation simplicity and compatibility with standard image-classification pipelines, we consider only square patches and square layouts of $\sqrt{Z} \times \sqrt{Z}$, leading to the following patch resolution:
\begin{equation}
    h_p = w_p = \frac{R}{\sqrt{Z}},
\end{equation}
where $R$ denotes the side length of the square input to the downstream model.

\par

Given a distillation budget of $\mathrm{IPC}$ images per class, the required number of patches per class is $N_p = Z \times \mathrm{IPC}$. Let $\mathcal{P}^{\star}_c = \{ \bm{p}^{\star}_i \mid i \in \mathcal{I}^{\mathrm{cand}}_c \}$ denote the selected patches of class $c$. We calculate their realism scores and retain the $N_p$ highest-scoring patches to form the key patch set $\mathcal{K}_c$ as follows:
\begin{equation}
\mathcal{K}_c = \operatorname{TopK} \left( \mathcal{P}^{\star}_c, S, N_p \right).
\label{eq:topk_selection}
\end{equation} 

The retained patches in $\mathcal{K}_c$ are sorted in descending order by their realism scores and sequentially partitioned into $\mathrm{IPC}$ groups $\{\mathcal{B}_{c,j}\}_{j=1}^{\mathrm{IPC}}$, each containing $Z$ patches. The patches in each group are concatenated in a fixed $\sqrt{Z} \times \sqrt{Z}$ layout to obtain the distilled image $\bm{x}^{\star}_{c,j}$ as follows: 
\begin{equation}
\bm{x}^{\star}_{c,j} = \operatorname{Aggregate}( \mathcal{B}_{c,j} ), 
\label{eq:patch_aggregation}
\end{equation}
where $\operatorname{Aggregate}$ denotes the aggregation process.

\par

Because an aggregated image $\bm{x}^{\star}_{c,j}$ can contain heterogeneous semantic content, we replace the original hard label with a soft label produced by the observer model following \citet{sun2024rded}. Specifically, we obtain $N_\mathrm{soft}$ random crops $\{ \bm{x}^{\star(n)}_{c,j} \} _{n=1} ^{N_\mathrm{soft}}$ from $\bm{x}^{\star}_{c,j}$ and use them to capture the characteristics of different image regions. The soft label $\bm{q}_{c,j}$ is defined as the average class-probability distribution predicted by the observer model $\phi_{\theta_T}$ as follows:
\begin{equation}
\bm{q}_{c,j}=\frac {1} {N_\mathrm{soft}} \sum_{n=1} ^{N_\mathrm{soft}} \phi _{\theta_T} \bigl( \bm{x} ^{\star(n)} _{c,j} \bigr).
\label{eq:soft_label}
\end{equation}

\par

With these efforts, the final distilled dataset is obtained as $\mathcal{D}^{\star} = \bigcup_{c=1} ^{C} \{ (\bm{x} ^{\star} _{c,j}, \bm{q}_{c,j}) \} _{j=1} ^{\mathrm{IPC}}$. 

\section{Experiments}
We evaluate the proposed method from several complementary perspectives. We first compare it with representative optimization-based and selection-based dataset distillation methods on standard image-classification benchmarks and then examine its scalability to a dataset with more categories. We further analyze the effect of category-specific routing threshold $\mathcal{T}_c$, patches per image $Z$, transformation strategies, and routing strategies. Finally, we compare the time required for foreground extraction and dataset synthesis. 

\begin{table*}[t]
\centering
\footnotesize
\caption{Comparison of test accuracy (\%) on standard benchmarks with baseline methods. Results are reported as mean $\pm$ standard deviation. The best results are highlighted in bold.}
\label{tab:main}
\begin{tabular}{c|c|ccc|ccc|ccc}
\hline

\multicolumn{2}{c|}{Dataset} 
& \multicolumn{3}{c|}{ImageNette} 
& \multicolumn{3}{c|}{ImageWoof} 
& \multicolumn{3}{c}{CIFAR-10} 
\\

\multicolumn{2}{c|}{IPC} 
& 1 & 10 & 50 
& 1 & 10 & 50 
& 1 & 10 & 50 
\\
\hline

\multirow{8}{*}{ResNet-18} 
& Random
& $18.9 \pm 3.2$ & $40.3 \pm 2.7$ & $62.7 \pm 1.7$ 
& $14.6 \pm 2.1$ & $23.1 \pm 1.3$ & $38.4 \pm 1.7$

& $15.4 \pm 1.3$ & $28.9 \pm 2.1$ & $43.3 \pm 2.2$ 
\\

& CDA
& $25.4 \pm 0.6$ & $54.6 \pm 0.4$ & $77.8 \pm 0.3$
& $14.6 \pm 0.6$ & $25.7 \pm 0.5$ & $59.7 \pm 0.5$
& $16.4 \pm 0.6$ & $30.6 \pm 0.6$ & $54.5 \pm 0.7$
\\

& G-VBSM

& $28.9 \pm 0.6$ & $61.6 \pm 0.4$ & $81.4 \pm 0.7$ 
& $14.4 \pm 0.4$ & $34.5 \pm 0.5$ & $65.5 \pm 0.5$ 
& $17.5 \pm 0.7$ & $31.5 \pm 0.4$ & $55.6 \pm 0.4$
\\

& DWA

& $29.7 \pm 0.9$ & $64.3 \pm 0.4$ & $83.2 \pm 0.5$ 
& $16.5 \pm 0.5$ & $36.1 \pm 0.5$ & $67.8 \pm 0.7$ 
& $18.3 \pm 0.3$ & $33.1 \pm 0.4$ & $59.9 \pm 0.4$ \\


& SRe$^2$L
& $19.1 \pm 1.1$ & $29.4 \pm 3.0$ & $40.9 \pm 0.3$ 
& $13.3 \pm 0.5$ & $20.2 \pm 0.2$ & $23.3 \pm 0.3$ 
& $16.6 \pm 0.9$ & $29.3 \pm 0.5$ & $45.0 \pm 0.7$ 
\\
& DM
& $21.4\pm0.3$ & $54.1\pm1.3$ & $68.6\pm0.7$ 
& $11.2\pm1.4$ & $23.2\pm1.4$ & $34.8\pm0.9$ 
& $14.2\pm0.3$ & \bm{$38.6\pm0.3$} & $55.6\pm0.5$ 
\\

& RDED
& $28.8 \pm 2.9$ & $65.4 \pm 2.0$ & $85.3 \pm 1.0$ 
& $20.9 \pm 0.8$ & $46.0 \pm 1.0$ & $68.7 \pm 0.9$
& $20.3 \pm 1.4$ & $36.4 \pm 1.5$ & $59.8 \pm 1.0$ 
\\

& Ours
& \bm{$29.8 \pm 3.0$} & \bm{$66.6 \pm 1.1$} & \bm{$85.8 \pm 1.2$}
& \bm{$22.4 \pm 0.5$} & \bm{$47.7 \pm 1.5$} & \bm{$69.7 \pm 0.6$} 
& \bm{$20.5 \pm 1.1$} & \bm{$38.6 \pm 0.9$} & \bm{$64.0 \pm 0.9$} 
\\
\hline

\multirow{6}{*}{ConvNet} 
& Random
& $19.3 \pm 4.4$ & $39.3 \pm 2.5$ & $60.9 \pm 1.9$ 
& $14.1 \pm 1.0$ & $22.5 \pm 1.6$ & $33.1 \pm 1.2$ 
& $18.2 \pm 2.4$ & $31.2 \pm 0.1$ & $46.6 \pm 1.7$ 
\\

& K-Center
& $19.5 \pm 4.8$ & $20.7 \pm 2.6$ & $33.5 \pm 1.5$ 
& $14.1 \pm 0.9$ & $15.0 \pm 0.7$ & $18.9 \pm 1.0$ 
& $18.2 \pm 2.4$ & $24.5 \pm 3.1$ & $33.3 \pm 0.7$ 
\\

& Herding
& $23.1 \pm 0.5$ & $46.1 \pm 1.3$ & $63.3 \pm 0.6$ 
& $15.9 \pm 0.6$ & $20.3 \pm 0.9$ & $33.1 \pm 1.0$
& $19.8 \pm 0.5$ & $33.7 \pm 0.2$ & $46.7 \pm 0.4$ 
\\

& DM
& $28.3\pm0.9$ & $58.4\pm1.3$ & $66.1\pm2.0$ 
& $20.1\pm0.5$ & $30.6\pm0.5$ & $38.1\pm1.3$ 
& $25.8\pm0.4$ & $49.5\pm0.5$ & $62.4\pm0.1$ 
\\

& RDED
& $28.3 \pm 2.5$ & $62.4 \pm 1.6$ & $83.0\pm 1.3$ 
& $20.9 \pm 1.2$ & $41.2 \pm 1.2$ & $61.3 \pm 0.8$ 
& $23.4 \pm 2.3$ & $49.2 \pm 1.6$ & $66.5 \pm 0.5$
\\

& Ours
& \bm{$30.9\pm 2.2$} & \bm{$64.5 \pm 1.6$} & \bm{$83.5 \pm 0.7$}
& \bm{$22.3 \pm 1.8$} & \bm{$41.8 \pm 1.1$} & \bm{$62.1 \pm 0.9$} 
& \bm{$26.3 \pm 1.4$} & \bm{$50.6 \pm 1.1$} & \bm{$67.4 \pm 0.6$}
\\
\hline

\end{tabular}
\end{table*}
\begin{table}[t]
 \caption{Comparison of test accuracy (\%) on CIFAR-100 with other baseline methods. Results are reported as mean $\pm$ standard deviation. The best results are highlighted in bold.}
 \label{tab:scale}
 \begin{tabular}{c|c|ccc}
 \hline
 \multicolumn{2}{c|}{Dataset} 
 & \multicolumn{3}{c}{CIFAR-100} 
 \\
 
 \multicolumn{2}{c|}{IPC} 
 & 1 & 10 & 50 
 \\
 \hline
 
 \multirow{3}{*}{ResNet-18} 
 & SRe$^2$L
 & $6.6 \pm 0.2$ & $27.0 \pm 0.4$ & $50.2 \pm 0.4$ \\
 & RDED
 & $11.4 \pm 0.8$ & $44.6 \pm 0.5$ & $63.4 \pm 0.2$ 
 \\
 
 & Ours
 & \bm{$12.8 \pm 0.8$} & \bm{$47.9 \pm 0.7$} & \bm{$64.6 \pm 0.2$} 
 \\
 \hline
 
 \multirow{6}{*}{ConvNet} 
 & MTT
 & $24.3 \pm 0.3$ & $40.1 \pm 0.4$ & $47.7 \pm 0.2$ 
 \\
 
 & IDM
 & $20.1 \pm 0.3$ & $45.1 \pm 0.1$ & $50.0 \pm 0.2$ 
 \\
 
 & TESLA
 & $24.8 \pm 0.5$ & $41.7 \pm 0.3$ & $47.9 \pm 0.3$ 
\\

 & DATM
 & \bm{$27.9 \pm 0.2$} & $47.2 \pm 0.4$ & $55.0 \pm 0.2$ 
 \\
 
 & RDED
 & $20.9 \pm 0.8$ & $48.3 \pm 0.5$ & \bm{$57.5 \pm 0.2$}
 \\
 
 & Ours
 & $21.5 \pm 0.7$ & \bm{$48.9 \pm 0.3$} & \bm{$57.5 \pm 0.2$}
 \\
 \hline

 \end{tabular}
\end{table}

\subsection{Overall Experimental Setup}
We evaluate our method on three image-classification benchmarks including CIFAR-10, ImageNette, and ImageWoof, and further extend it to CIFAR-100. CIFAR-10 and CIFAR-100 contain $32 \times 32$ color images from 10 and 100 classes, respectively. ImageNette and ImageWoof are 10-class ImageNet subsets, each with approximately 13,000 high-resolution images. ImageNette contains visually distinct categories, whereas ImageWoof focuses on fine-grained dog breeds.

\par

Following common practice in dataset distillation, we use ConvNet and ResNet-18 as evaluation backbones. For each dataset and backbone, we report results at $\text{IPC} \in \{1, 10, 50\}$, where IPC denotes the number of distilled images per class. Unless otherwise stated, target models are trained from scratch on the distilled dataset $\mathcal{D}^{\star}$ with standard supervised learning for $300$ epochs and evaluated on the untouched test set for $3$ times.

\par

For each class, both RDED and our method first uniformly sample the same candidate subset from the original training data. Unless otherwise stated, the observer model has the same architecture as the validation model. The candidate set contains $N_{\mathrm{cand}} = 300$ images per class. The routing rate is $\rho=0.3$. For the candidate-crop path, the cropping follows the settings in RDED, with $K_\mathrm{crop}=5$, RRC scale of $(0.08, 1.0)$, and RRC ratio of $(1.0, 1.0)$. For the direct transformation path, the default transformation strategy is Squash. The number of patches per image is set to $Z=4$ for ImageWoof, and $Z=1$ for ImageNette, CIFAR-10, and CIFAR-100. 

\vspace{-6pt}
\subsection{Analysis of Foreground Occupancy Distributions}
Foreground occupancy is the routing cue in the proposed method. Before reporting benchmark results, we examine the class-wise distributions of foreground occupancy ratios for ImageNette, ImageWoof, and CIFAR-10. As shown in Fig.~\ref{fig:distribution}, each bar counts the images of a class whose $r_i$ falls within the corresponding bin, and Images without valid foreground masks are excluded from the bars. The dashed line marks the routing threshold $\mathcal{T}_c$. Images below the threshold (blue areas) are routed to the candidate-crop selection path, denoted as Crop, whereas the remaining images (orange areas) are routed to the direct transformation path, denoted as Transform. Detailed image numbers are listed in the spare area. The distributions vary substantially across both datasets and classes, indicating that a single global threshold would not adequately characterize their different image compositions.

\par

Specifically, ImageNette contains several strongly background-dominated classes, such as Chain Saw, Golf Ball, and Parachute, resulting in relatively small thresholds. In comparison, the thresholds of ImageWoof are distributed within a narrower range, reflecting relatively consistent yet still class-dependent object scales across its fine-grained dog breeds. CIFAR-10 instead includes classes concentrated in higher-occupancy regions. These observations demonstrate that foreground occupancy depends on both dataset characteristics and composition, leading to class-specific settings.

\par

The distributions motivate the proposed category-specific routing thresholds by using a routing ratio to decide the relative dominance of an image. Such thresholds route approximately the same proportion of images to the candidate-crop selection path for each class, preventing the routing decision from being disproportionately affected by a fixed threshold when processing classes with inherently different foreground occupancy ratios. We also examine the influence of the thresholds in Sec.~\ref{sec:threshold}.

\subsection{Comparison on Standard Benchmarks}
We compare our method with a range of representative dataset distillation baselines on ImageNette, ImageWoof, and CIFAR-10. For the ResNet-18 backbone, the baselines include Random, CDA~\citep{yin2024dataset}, G-VBSM~\citep{shao2024generalized}, DWA~\citep{du2024diversity}, SRe$^2$L~\citep{yin2023sre2l}, DM~\citep{zhao2023dm}, and RDED~\citep{sun2024rded}. For the ConvNet backbone, we compare against Random, K-Center~\citep{sener2017kcenter}, Herding~\citep{Welling2009Herding}, DM~\citep{zhao2023dm}, and RDED~\citep{sun2024rded}.

\par

The quantitative results are reported in Table~\ref{tab:main}. Our method consistently improves upon RDED on the standard benchmarks and achieves competitive or leading performance compared with the other evaluated baselines across different network architectures and distillation budgets.

\begin{figure*}[t]
    \centering
    \subfigure[ImageNette]{
        \includegraphics[width=5.3cm]{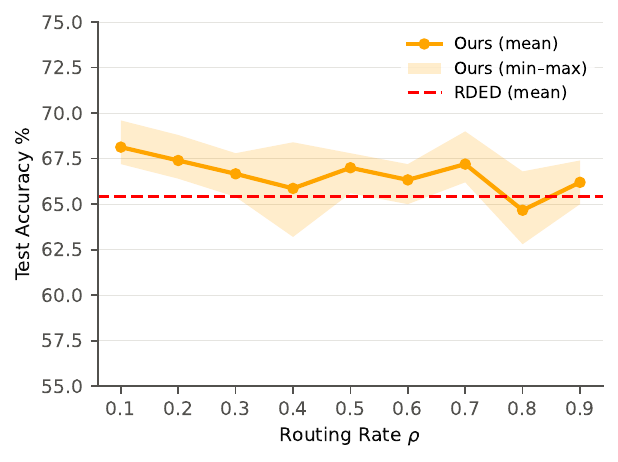}
    }
    \subfigure[ImageWoof]{
        \includegraphics[width=5.3cm]{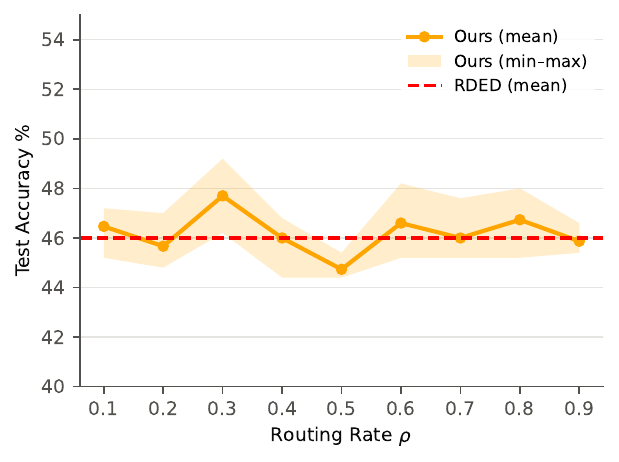}
    }
    \subfigure[CIFAR-10]{
        \includegraphics[width=5.3cm]{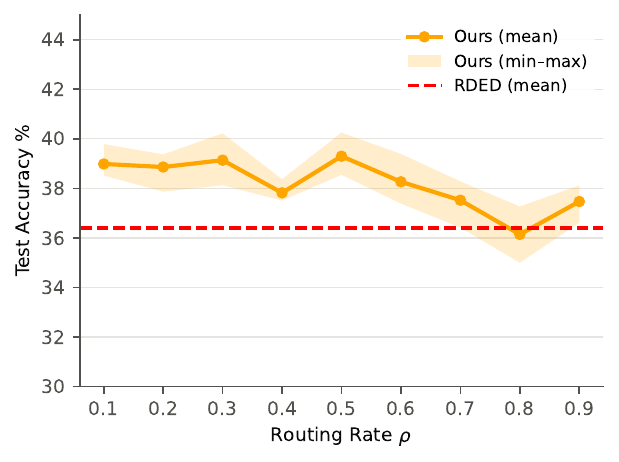}
    }
    \caption{Effect of the category-specific routing threshold $\mathcal{T}_c$ on ImageNette, ImageWoof, and CIFAR-10 using $\mathrm{IPC}=10$ with ResNet-18. The horizontal axis denotes the routing rate $\rho$ that defines $\mathcal{T}_c$, and the vertical axis denotes test accuracy. The shadow denotes the minimum and maximum values. The dashed line shows the baseline RDED.
    }
    \label{fig:rate}
\end{figure*}
\begin{figure*}[t]
    \centering
    \subfigure[ImageNette]{
        \includegraphics[width=5.3cm]{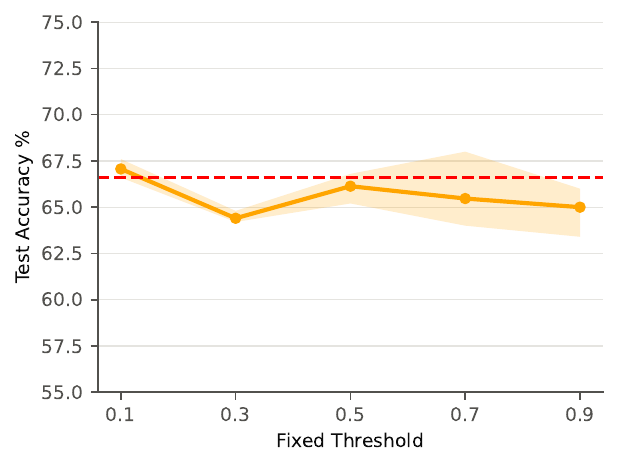}
    }
    \subfigure[ImageWoof]{
        \includegraphics[width=5.3cm]{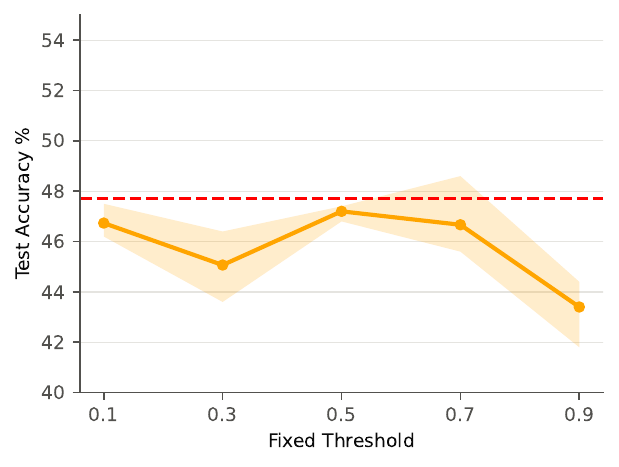}
    }
    \subfigure[CIFAR-10]{
        \includegraphics[width=5.3cm]{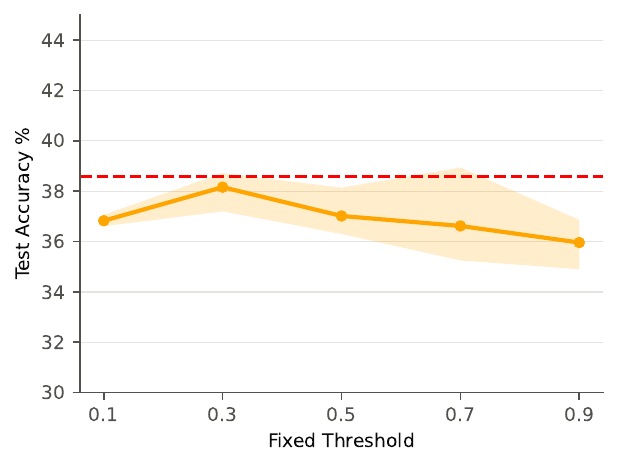}
    }
    \caption{Analysis on fixed thresholds. The experiment is conducted on ImageNette, ImageWoof, and CIFAR-10 using $\mathrm{IPC}=10$ with ResNet-18. The horizontal axis denotes the fixed threshold, and the vertical axis denotes test accuracy. The shadow denotes the minimum and maximum values. The dashed line shows the test accuracy using the routing rate $\rho=0.3$.
    }
    \label{fig:fixed_th}
\end{figure*}

\subsection{Scalability to a Larger-Scale Dataset}
We further evaluate the scalability of our method on CIFAR-100, which has a substantially larger number of categories. We consider ResNet-18 and ConvNet backbones under $\text{IPC} \in \{1, 10, 50\}$. For the ResNet-18 backbone, we compare our method with SRe$^2$L~\citep{yin2023sre2l} and RDED~\citep{sun2024rded}. For the ConvNet backbone, we include MTT~\citep{cazenavette2022dataset}, IDM~\citep{zhao2023idm}, TESLA~\citep{cui2023scaling}, DATM~\citep{guo2024datm}, and RDED~\citep{sun2024rded} as baselines.

\par

As shown in Table~\ref{tab:scale}, our method achieves consistent improvements or comparable performance relative to RDED across various settings. It also achieves the best performance at $\mathrm{IPC}=10$ and $\mathrm{IPC}=50$. These results indicate that foreground-aware patch routing scales to a more complex classification setting. By generating patches according to category-relative occupancy, our method better retains object details that may be important for discrimination between a large number of categories.

\subsection{Effect of the Routing Threshold \texorpdfstring{$\mathcal{T}_c$}{Tc}}
\label{sec:threshold}
The dual-path selection strategy is controlled by the category-specific routing thresholds $\{\mathcal{T}_c\}_{c=1}^{C}$, which are derived from the class-wise foreground-occupancy distributions according to the routing rate $\rho$. As a result, we vary and report the shared $\rho$ rather than the individual threshold values, with a single $\rho$ determining a unique set of thresholds. A small $\rho$ favors direct transformation and may retain unnecessary background content, whereas a large $\rho$ routes more images to the candidate-crop selection path, increasing the risk of removing important foreground structures. To examine the sensitivity of the method to this trade-off, we vary $\rho$ from $0.1$ to $0.9$ in increments of $0.1$ and evaluate the resulting threshold sets on ImageNette, ImageWoof, and CIFAR-10 at $\mathrm{IPC}=10$ using ResNet-18.

\par

As shown in Fig.~\ref{fig:rate}, the effect of the thresholds is non-monotonic and varies across datasets. ImageNette performs better at relatively small $\rho$s, whereas ImageWoof and CIFAR-10 achieve their peak performance at larger $\rho$s. Nevertheless, an excessively large $\rho$ is unfavorable across all three datasets. This result reflects the trade-off between foreground concentration through the candidate-crop selection path and foreground preservation through the direct transformation path.

\par

The foreground-occupancy distributions in Fig.~\ref{fig:distribution} provide a possible explanation for this dataset-dependent behavior. Compared with ImageWoof and CIFAR-10, ImageNette contains more classes whose occupancy distributions are concentrated at lower regions. In these images, foreground objects occupy only a small fraction of the image. Although the candidate-crop selection path evaluates multiple crops using the realism score, it remains difficult to guarantee that the selected crop contains the complete foreground object. Such risk may diminish the benefit from foreground concentration through cropping. Increasing $\rho$ routes a larger proportion of such images to the crop path, thereby amplifying the associated risks, shifting the balance point of ImageNette toward smaller routing rates. In contrast, ImageWoof and CIFAR-10 contain more classes with moderate or high average foreground occupancy ratios. A randomly sampled crop is consequently more likely to retain sufficient foreground evidence. With reduced risk of suboptimal cropping, the proposed method favors a higher proportion of candidate-crop selection paths in ImageWoof and CIFAR-10 relative to ImageNette.

\par

We also compare the proposed strategy with a fixed-threshold alternative, where a same absolute threshold is applied to all classes. As shown in Fig.~\ref{fig:fixed_th}, we evaluate threshold values from $0.1$ to $0.9$ in increments of $0.2$ under the same setting and compare the accuracy with that obtained using $\rho=0.3$. The optimal threshold value varies across datasets due to the variation in dataset-level foreground-occupancy distributions. In comparison, using category-specific thresholds achieves better or comparable performance across the evaluated datasets. Although a carefully selected fixed threshold can occasionally achieve comparable or slightly higher accuracy, the proposed quantile-based strategy reduces the reliance on manually specified absolute occupancy thresholds while maintaining competitive performance.

\par

Overall, these results suggest that the appropriate balance between the two paths depends on the dataset-level foreground-occupancy distribution. Considering the performance across all evaluated datasets, we adopt $\rho=0.3$ as a common default without dataset-specific tuning.

\begin{table}[t]
    \centering
    \caption{Ablation study for $Z$. Experiments are conducted on CIFAR-10, ImageNette, and ImageWoof under $\mathrm{IPC}=10$ using ResNet-18. Results are reported as mean $\pm$ standard deviation. The best results are highlighted in bold.}
    \label{tab:z}
    \begin{tabular}{c|ccc}
        \hline
        Dataset & $Z=1$ & $Z=4$ & $Z=16$ \\
        \hline

        ImageNette 
        & \bm{$66.6 \pm 1.1 $}
        & $63.5 \pm 1.4$
        & $57.1 \pm 2.0$ 
        \\
        
        ImageWoof 
        & $42.7 \pm 2.1 $
        & \bm{$47.7 \pm 1.5$} 
        & $35.3 \pm 1.1$ 
        \\

        CIFAR-10
        & \bm{$38.6 \pm 0.9 $}
        & $36.1 \pm 1.9$ 
        & $25.1 \pm 3.0 $
        \\
        
        \hline
    \end{tabular}
\end{table}
\subsection{Effect of Patches per Image \texorpdfstring{$Z$}{Z}}
\label{sec:z}
In our framework, $Z$ specifies the number of patches aggregated into a single distilled image. When $Z$ is small, each cropped patch covers a larger region of the source image, while each transformed patch retains a clearer view of the complete image. Although the distilled image contains fewer constituent patches, each patch preserves more foreground information and discriminative detail. Increasing $Z$ allows information from more source images to be incorporated, thereby improving information diversity. However, cropped patches occupy fewer pixels, and transformed patches undergo stronger compression. Both effects may lead to the loss of foreground details. Overall, $Z$ controls a trade-off between the spatial fidelity of individual patches and the information diversity of the distilled image.

\par

As shown in Table~\ref{tab:z}, the optimal value of $Z$ depends on the characteristics of the dataset. For CIFAR-10, whose images have a native resolution of only $32\times32$, the highest accuracy is achieved at $Z=1$. Increasing $Z$ to $4$ already results in patches of substantially low resolution and reduced visual quality. For the ImageNet subsets, whose images are typically processed at a resolution of $224\times224$, the optimal value depends on the visual characteristics of each subset. ImageNette is designed as a comparatively easy classification dataset and contains categories that are visually distinct from one another. It achieves its highest accuracy at $Z=1$, suggesting that a single high-quality patch generally provides sufficient information for recognition. In contrast, ImageWoof consists of fine-grained dog breeds whose discrimination requires more subtle and diverse visual cues. It achieves its highest accuracy at $Z=4$, indicating that combining complementary regions from multiple source images is beneficial in this fine-grained setting. Nevertheless, $Z=16$ substantially degrades performance on all three datasets, demonstrating that excessive patch aggregation is detrimental due to the severe loss of spatial fidelity within each patch. Based on these results, we use $Z=4$ for ImageWoof, and $Z=1$ for ImageNette, CIFAR-10, and CIFAR-100.

\begin{figure*}[t]
        \centering
        \includegraphics[width=17cm]{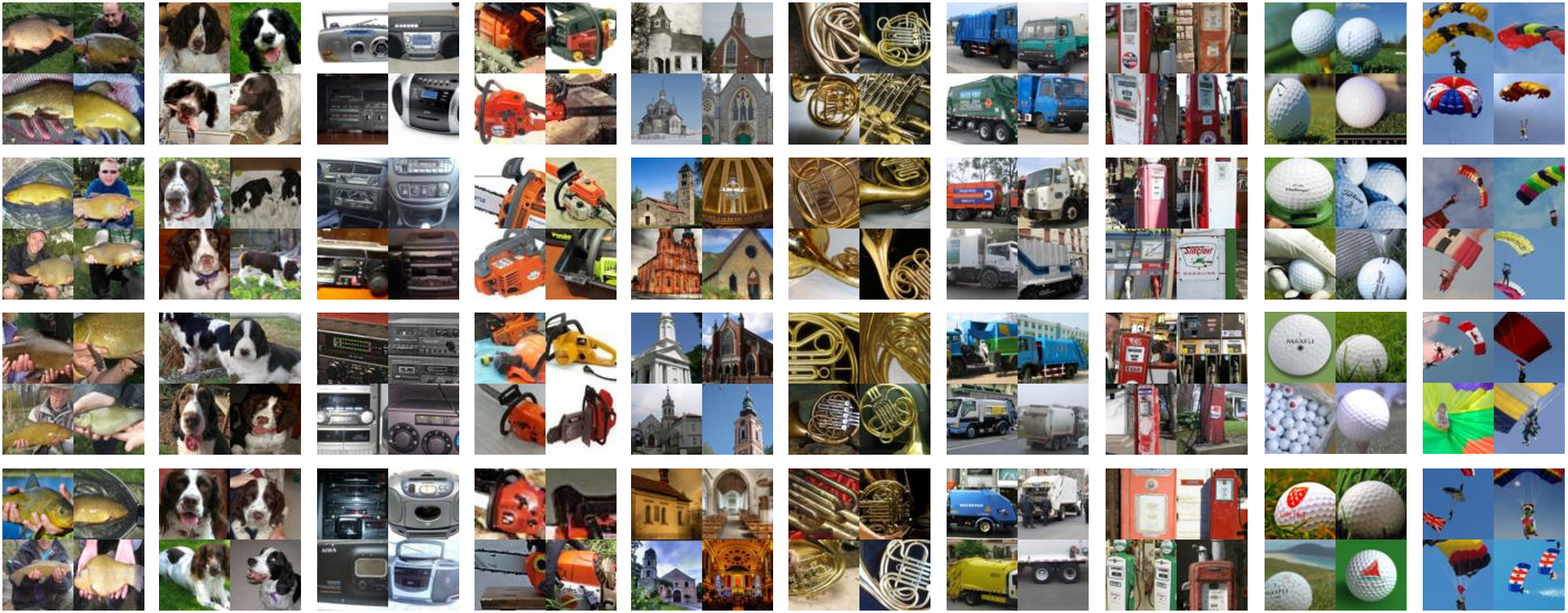}
        \caption{Visualization results of distilled images for ImageNette.}
        \label{fig:nette}

        \vspace{0.4cm}

        \includegraphics[width=17cm]{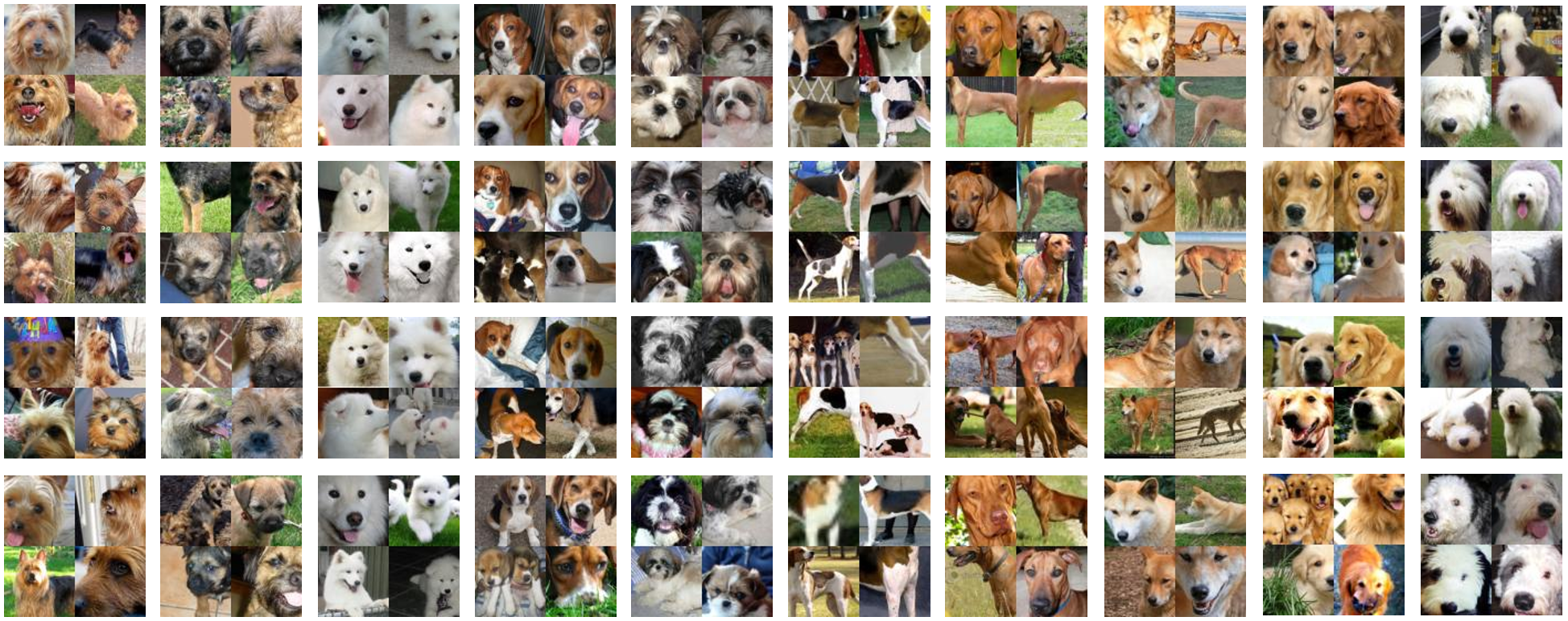}
        \caption{Visualization results of distilled images for ImageWoof.}
        \label{fig:woof}
\end{figure*}

\begin{table}[t]
\centering
\caption{Comparison of test accuracy (\%) between different transformation strategies on ImageWoof using ResNet-18. Results are reported as mean $\pm$ standard deviation. The best results are highlighted in bold.}
\label{tab:transform}
\begin{tabular}{l|ccc}
\hline
Transformation & IPC = 1 & IPC = 10 & IPC = 50 
\\
\hline

Squash
& \bm{$22.4 \pm 0.5$} & \bm{$47.7 \pm 1.5$} & $69.7 \pm 0.6$ 
\\

CenterCrop 
& $21.3 \pm 1.3$ & $46.5 \pm 2.2$ & $69.6 \pm 0.9$ 
\\

RandomSquare
& $21.3 \pm 1.6$ & $47.2 \pm 2.4$ & $69.3 \pm 1.2$
\\

RRC
& $19.8 \pm 0.8$ & $47.1 \pm 2.5$ & \bm{$70.0 \pm 0.2$}
\\
\hline
\end{tabular}
\end{table}

\subsection{Ablation on Transformation Strategies}
\label{sec:transform}
For the direct transformation path, we squash the image to preserve as much foreground information as possible. However, squashing may alter the aspect ratio of the original image, resulting in geometrically distorted and less realistic foreground objects. To further examine its effect on downstream accuracy, we introduce several alternative transformation strategies, including \emph{CenterCrop}, \emph{RandomSquare}, and \emph{RandomResizedCrop}. Specifically, CenterCrop extracts a square region centered in the image whose side length equals the shorter dimension of the image. \emph{RandomSquare} extracts a square region of the same size but at a random position. \emph{RandomResizedCrop} samples regions with a scale in $[0.5,1.0]$ and an aspect ratio in $[3/4,4/3]$. For the two stochastic operators, RandomSquare and RRC, we generate five patches from each source image and resize the one with the highest realism score to the target resolution as the selected patch.

\par

We compare the performance of different strategies on ImageWoof using ResNet-18 under $\mathrm{IPC} \in \{1, 10, 50\}$. As shown in Table~\ref{tab:transform}, Squash achieves the highest accuracy at $\mathrm{IPC}=1$ and $\mathrm{IPC}=10$, possibly because the subsequent realism-based ranking makes severely distorted or poorly recognizable squashed patches less likely to enter the final distilled set under low distillation budgets. In this regime, preserving the complete foreground and global structure is therefore more beneficial than avoiding geometric distortion. CenterCrop and RandomSquare do not consistently improve performance, suggesting that avoiding squash-induced distortion does not necessarily compensate for the loss of foreground information and global structure due to cropping. RRC enhances diversity at the expense of increased information loss, achieving the best performance at $\mathrm{IPC}=50$. These results demonstrate that the balance between visual fidelity and foreground preservation varies across distillation budgets. Squash provides the most stable performance across IPC settings and is therefore used as the default transformation strategy.

\begin{figure*}[t]
        \centering
        \includegraphics[width=17cm]{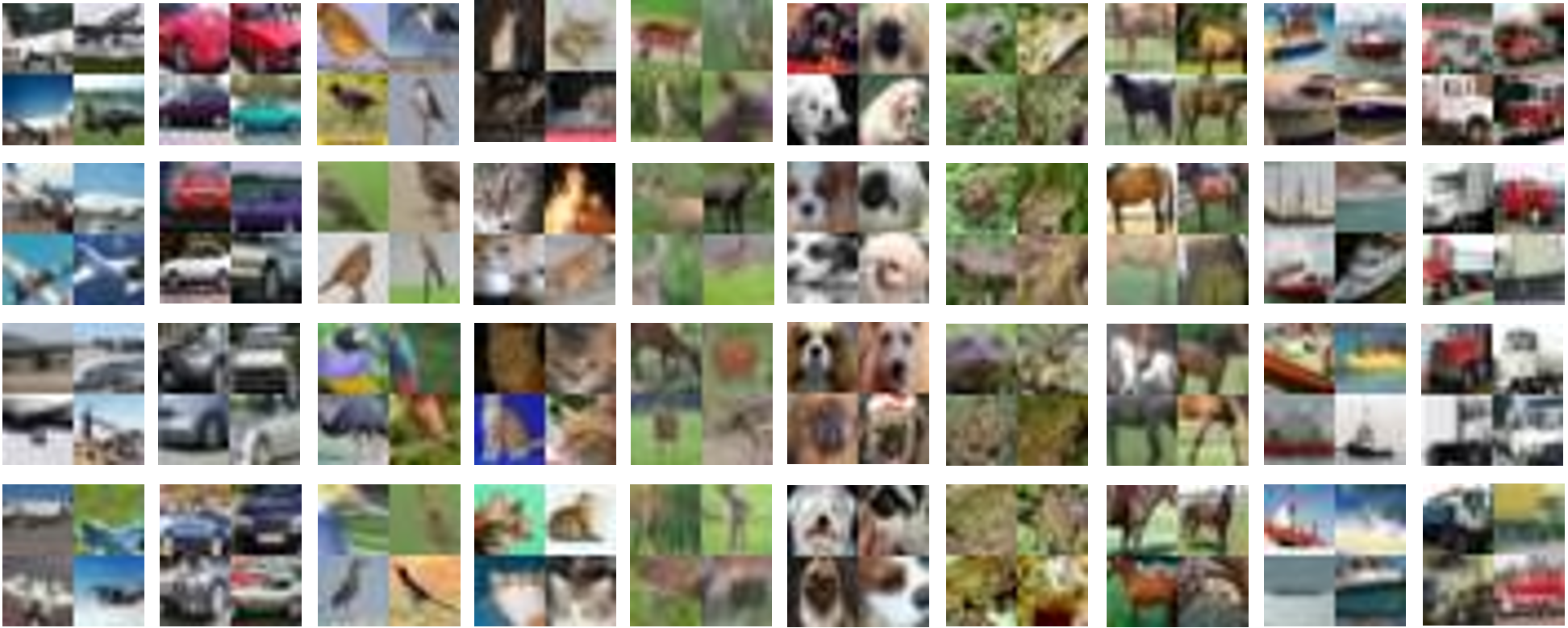}
        \caption{Visualization results of distilled images for CIFAR-10.}
        \label{fig:cifar10}

        \vspace{0.4cm}

        \includegraphics[width=17cm]{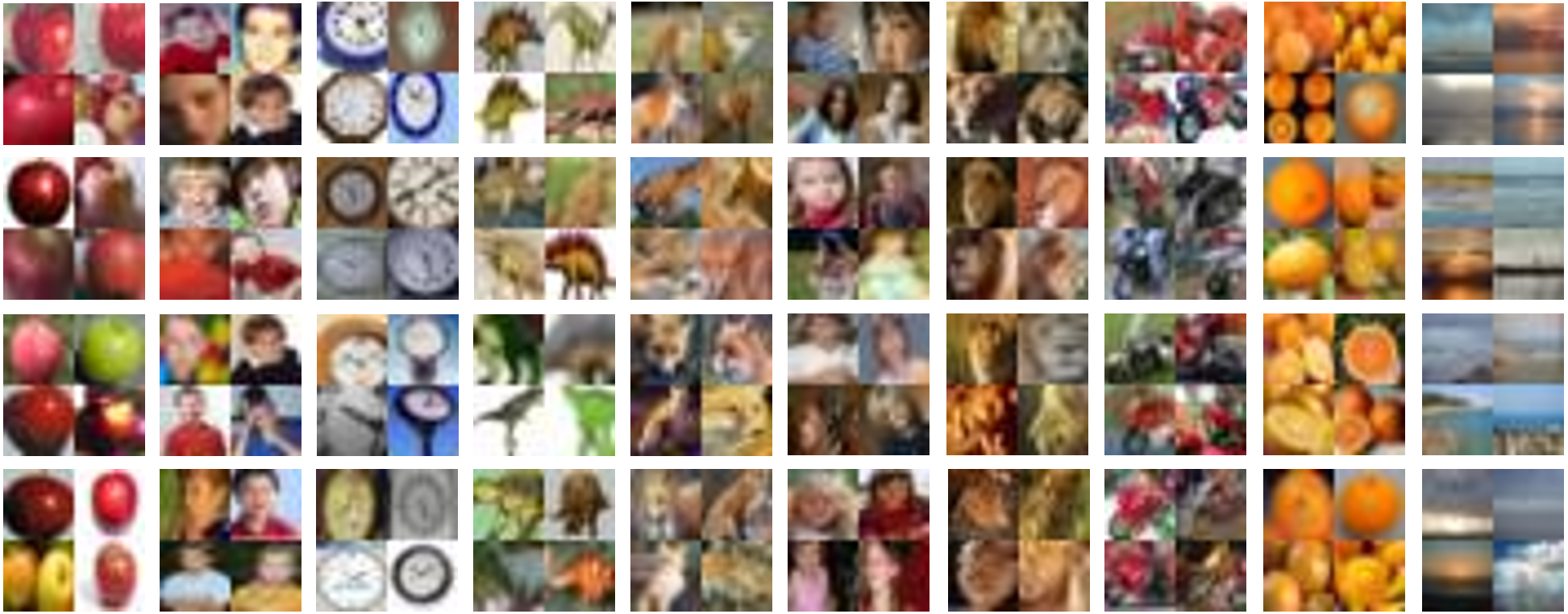}
        \caption{Visualization results of distilled images for CIFAR-100.}
        \label{fig:cifar100}
\end{figure*}

\subsection{Ablation on Routing Strategies}
To further validate the effectiveness of foreground-aware dual-path routing, we compare our method with several alternative routing strategies on ImageWoof using ResNet-18. Specifically, we consider routing all candidate images to the candidate-crop selection path, as performed in RDED (\emph{AllCrop}); routing all images to the direct transformation path (\emph{AllTransform}); and randomly routing the same number of images to the two paths as in our method (\emph{RandomRoute}). This setting allows us to determine whether the improvement arises merely from combining cropping and transformation.
\begin{table}[t]
\centering
\caption{Comparison of test accuracy (\%) between different routing strategies on ImageWoof using ResNet-18. Results are reported as mean $\pm$ standard deviation. The best results are highlighted in bold.}
\label{tab:route}
\begin{tabular}{l|ccc}
\hline
Routing Strategy & IPC = 1 & IPC = 10 & IPC = 50 
\\
\hline

AllCrop
& $20.9 \pm 0.8$ & $46.0 \pm 1.0$ & $68.7 \pm 0.9$ 
\\

AllTransform 
& $20.5 \pm 0.6$ & $44.5 \pm 0.5$ & $68.9 \pm 0.3$ 
\\

RandomRoute 
& $20.2 \pm 0.2$ & $45.4 \pm 0.9$ & $67.8 \pm 0.4$ 
\\

Ours 
& \bm{$22.4 \pm 0.5$} & \bm{$47.7 \pm 1.5$} & \bm{$69.7 \pm 0.6$}
\\

\hline
\end{tabular}
\end{table}

\begin{table}[t]
    \centering
    \caption{Comparison of time cost between 
    foreground extraction (Foreground), RDED dataset synthesis (RDED), dual-path dataset synthesis (Ours), and downstream validation (Val) using ResNet-18.}
    \label{tab:cost}
    \setlength{\tabcolsep}{5pt}
    \begin{tabular}{cc|cccc}
        \hline
        Dataset & IPC & Foreground & RDED & Ours & Val\\
        \hline
        \multirow{3}{*}{ImageNette}
        & 1  
        & \multirow{3}{*}{\makecell{28.3\\min}}
        & 18.1 s & 15.2 s & 18.3 s \\
        & 10 & & 6.8 s & 4.8 s & 88.6 s\\
        & 50 & & 6.9 s & 4.9 s & 177.4 s\\
        \hline
        
        \multirow{3}{*}{ImageWoof}
        
        & 1  
        & \multirow{3}{*}{\makecell{27.6\\min}}
        & 17.1 s & 14.0 s & 15.8 s \\
        & 10 & & 6.2 s & 4.6 s & 86.7 s \\
        & 50 & & 6.4 s & 4.8 s & 176.1 s \\
        \hline
        
        \multirow{3}{*}{CIFAR-10}
        
        & 1 
        & \multirow{3}{*}{\makecell{95.4\\min}}
        & 3.4 s & 2.4 s & 24.4 s \\
        & 10 & & 1.2 s & 1.2 s & 43.4 s \\
        & 50 & & 1.3 s & 1.5 s & 62.0 s \\
        \hline
    \end{tabular}
\end{table}

\par

As shown in Table~\ref{tab:route}, our method achieves the best performance in all evaluated settings. Among the alternatives, AllCrop performs better at the lower IPC settings, whereas AllTransform achieves slightly higher accuracy at $\mathrm{IPC}=50$. This variation indicates that neither cropping nor direct transformation is uniformly suitable across different distillation budgets and image compositions. Meanwhile, RandomRoute does not consistently improve upon the fixed strategies and even yields the lowest accuracy in some settings. These results demonstrate that the improvement does not arise simply from combining cropping and transformation. Instead, it results from routing each image to an appropriate path according to its foreground occupancy, thereby balancing foreground concentration and preservation in a content-adaptive manner.

\vspace{-6pt}
\subsection{Computational Cost Analysis}
We compare the costs of foreground extraction (Foreground), RDED dataset synthesis (RDED), dual-path dataset synthesis (Ours), and downstream validation (Val). All experiments are conducted on a single NVIDIA H100 GPU. Foreground extraction is performed once for each dataset. Synthesis and validation times are averaged over three independent runs. 

\par

As shown in Table~\ref{tab:cost}, although foreground estimation introduces an additional prolonged processing time, the resulting masks and occupancy ratios can be cached and reused across all experiments concerned with the dataset. This one-time cost can be amortized over subsequent experiments. After foreground extraction, our method generally requires less synthesis time than RDED. This improvement can be attributed to the direct transformation path, which avoids repeated cropping and scoring in the candidate-crop selection path. The abnormally long synthesis time at $\mathrm{IPC}=1$ results from the default RDED implementation, which uses single-process data loading in this setting. We retain this setting for both methods to ensure a consistent comparison.  Overall, the proposed method retains the efficiency advantage of non-optimization-based distillation without introducing iterative optimization costs across each distillation.

\subsection{Visualization}
We present distilled images produced by our method with $Z=4$ on ImageNette, ImageWoof, CIFAR-10, and CIFAR-100 in Figs.~\ref{fig:nette}, \ref{fig:woof}, \ref{fig:cifar10}, and \ref{fig:cifar100}, respectively. For the low-resolution CIFAR-10 and CIFAR-100 datasets, dividing the images into four parts makes the constituent patches less visually detailed, resulting in the suboptimal performance discussed in Sec.~\ref{sec:z}. In contrast, on the high-resolution ImageNet subsets, the distilled images retain recognizable foreground characteristics, including object shapes, local textures, and structural details.

\par

The visualized images contain patches generated by both the candidate-crop selection path and the direct transformation path. Although both types of patches show concentrated foreground objects, cropped patches may omit peripheral regions due to randomly resized cropping, while transformed patches may exhibit geometric distortion caused by squash resizing. Our content-adaptive routing strategy balances the contributions of these two patch types to achieve a compromise between foreground concentration and structural preservation. Overall, the proposed method routes each source image to an appropriate patch-generation path and aggregates complementary information from multiple source images.

\section{Discussion}
The proposed method employs Grounded SAM2 to estimate foreground occupancy ratios for dual-path patch selection. First, inaccurate foreground masks may produce unreliable occupancy estimates and consequently lead to inappropriate routing decisions, particularly for images with occupancy ratios close to the routing threshold. Second, a scalar occupancy ratio cannot fully characterize foreground properties such as the number, shape, and spatial arrangement of objects. Moreover, the use of square patches and fixed aggregation layouts may also restrict the flexibility of distilled-image construction, especially for foreground objects with irregular geometries. Future work will investigate routing strategies that are more robust to segmentation errors, incorporate richer spatial statistics, and employ adaptive layouts tailored to foreground geometry.

\section{Conclusion}
In this work, we present a foreground-aware dataset distillation framework with content-adaptive dual-path patch selection. Instead of applying a fixed patch-construction strategy to all images, the proposed method uses foreground occupancy to route an image between the candidate-crop selection path and the direct transformation path. This design concentrates foreground information when cropping is beneficial while preserving object integrity by direct transformation when cropping would be destructive. Across CIFAR-10, CIFAR-100, ImageNette, and ImageWoof, the proposed method outperforms the baseline method RDED in nearly all evaluated settings. Analyses further show the effectiveness of category-specific routing thresholds, squash transformation, and content-adaptive routing. Although foreground estimation introduces an additional one-time cost, the extracted information can be reused across experiments, and the proposed synthesis procedure is generally faster than or comparable to RDED. These results demonstrate that incorporating foreground occupancy metrics into path routing during patch selection enhances the efficacy of non-optimization-based dataset distillation, indicating that lightweight structural cues can improve the informativeness of selected patches and the quality of the distilled dataset.

\section*{Acknowledgments}
This research was supported in part by JSPS KAKENHI Grant Numbers JP24K02942 and JP25K21218.

\bibliographystyle{elsarticle}
\bibliography{ref}

\end{document}